\documentclass[sigconf]{acmart}

\usepackage{graphicx} % Required for inserting images
\usepackage[ruled]{algorithm2e}
\usepackage{color}
\usepackage{enumitem}
\usepackage{float}
\usepackage{multirow}
\usepackage{booktabs}

\AtBeginDocument{%
  }

\setcopyright{acmlicensed}
\copyrightyear{2026}
\acmYear{2026}
\acmConference[SIGIR ’26]{the 49th International ACM SIGIR Conference on Research and Development in Information Retrieval}{July 20--24,
  2026}{Woodstock, NY}
\begin{document}

\title{Beyond Single Slot: Joint Optimization for Multi-Slot Guaranteed Display Advertising}


\author{Zhaoqi Zhang}
\affiliation{%
  \institution{Nanyang Technological University}
  \city{Singapore}
  \country{Singapore}
}
\affiliation{%
  \institution{Meituan}
  \city{Beijing}
  \country{China}
}

\email{zhaoqi001@e.ntu.edu.sg}

\author{Jiaming Deng}
\affiliation{%
  \institution{Meituan}
  \city{Beijing}
  \country{China}
}
\email{dengjiaming02@meituan.com}

\author{Miao Xie}
\affiliation{%
  \institution{China Agricultural University}
  \city{Beijing}
  \country{China}
}
\email{xiemiao@cau.edu.cn}

\author{Linyou Cai}
\affiliation{%
  \institution{Meituan}
  \city{Beijing}
  \country{China}
}
\email{cailinyou@meituan.com}

\author{Qianlong Xie}
\affiliation{%
  \institution{Meituan}
  \city{Beijing}
  \country{China}
}
\email{xieqianlong@meituan.com}

\author{Xingxing Wang}
\affiliation{%
  \institution{Meituan}
  \city{Beijing}
  \country{China}
}
\email{wangxingxing04@meituan.com}

\author{Siqiang Luo}
\affiliation{%
  \institution{Nanyang Technological University}
  \city{Singapore}
  \country{Singapore}
}
\email{siqiang.luo@ntu.edu.sg}

\author{Gao Cong}
\affiliation{%
  \institution{Nanyang Technological University}
  \city{Singapore}
  \country{Singapore}
}
\email{gaocong@ntu.edu.sg}

\renewcommand{\shortauthors}{Zhaoqi Zhang et al.}

\copyrightyear{2026}
\acmYear{2026}
\setcopyright{cc}
\setcctype{by}
\acmConference[SIGIR '26]{Proceedings of the 49th International ACM SIGIR Conference on Research and Development in Information Retrieval}{July 20--24, 2026}{Melbourne, VIC, Australia}
\acmBooktitle{Proceedings of the 49th International ACM SIGIR Conference on Research and Development in Information Retrieval (SIGIR '26), July 20--24, 2026, Melbourne, VIC, Australia}
\acmDOI{10.1145/3805712.3808398}
\acmISBN{979-8-4007-2599-9/2026/07}


\begin{abstract}
%Guaranteed Display (GD) advertising is crucial for platform monetization, yet existing methods often operate under a single-slot assumption, limiting their ability to optimize allocation across multi-slot page views. In this paper, we propose a novel joint optimization framework for multi-slot GD allocation, addressing key challenges such as slot-level redundancy, contract imbalance, and exposure concentration. Our approach formulates the allocation as an offline bipartite matching problem, enhanced with a contract roulette mechanism that ensures mutual exclusivity across slots and Page View constraints that control slot-level impression volumes. A distribution optimization algorithm efficiently solves the problem in large-scale settings. Extensive online A/B tests on Meituan demonstrates that our method significantly improves merchant ROI, platform revenue efficiency, and contract fulfillment robustness, demonstrating the applicability and effectiveness of our framework in real-world deployments.

Guaranteed display advertising is crucial for platform monetization, yet existing methods often operate under a single-slot assumption, limiting their ability to optimize allocation across multi-slot page views. In this paper, we propose a novel joint optimization framework for multi-slot GD allocation, addressing key challenges such as slot-level redundancy, contract imbalance, and exposure concentration. Our approach formulates the allocation as an offline bipartite matching problem with a contract roulette mechanism for slot exclusivity and Page View constraints for impression control, and incorporates a scalable allocation optimization algorithm for efficient large-scale deployment. Extensive online tests on the Meituan advertising platform demonstrate that our method significantly improves merchant ROI, platform revenue efficiency, and contract fulfillment robustness. Specifically, online A/B tests show a \textbf{28.99\%} increase in Average Revenue Per User under 70\% traffic, and DID analysis further indicates improved contract stability, demonstrating the strong applicability and effectiveness of our framework in real-world advertising deployments.
%Average Revenue Per User increased by \textbf{28.17\%}, and the Fulfillment Rate improved by \textbf{2.12\%}, demonstrating the strong applicability and effectiveness of our framework in real-world advertising deployments.

\end{abstract}





\maketitle

\section{Introduction}

In guaranteed display (GD) advertising, mainstream approaches are predominantly built upon a single-slot modeling assumption, where each ad contract is independently optimized for a specific slot to maximize engagement metrics such as Click-Through Rate(CTR) and Payment Conversion Rate(CVR)~\cite{hojjat2014delivering, fang2019large, lei2020multi}. While effective in early-stage settings with simpler ad placements, such approaches are increasingly inadequate for modern platforms such as Meituan and Taobao, where a single page may contain multiple ad slots, each page view can trigger simultaneous exposures across several slots, rendering independent slot-wise optimization insufficient for capturing complex allocation dynamics.

Although recent works~\cite{mao2023end, li2024bi} introduce coarse supply-side constraints to prevent over-delivery, they lack fine-grained slot-level control, allowing a few high-priority contracts to dominate premium positions. Moreover, most systems rely on online greedy allocation~\cite{dai2024percentile, lei2025generative}, which makes slot-wise decisions in isolation. Consequently, current methods remain insufficient to ensure fairness, exposure diversity, and stable delivery across multiple slots.

Despite their effectiveness in improving fulfillment and click performance, mainstream GD methods such as AUAF~\cite{cheng2022adaptive} exhibit critical limitations in real-world multi-slot environments: \textbf{(1) Lack of coordination across ad slots:} Most existing methods adopt a local modeling approach based on a single-slot view, where each optimization step considers only the allocation between one ad slot and several contracts. This localized strategy may lead to the overuse of popular slots and the underuse of others, reducing overall delivery efficiency. \textbf{(2) No upper limit on contract impressions:} While most methods ensure minimum delivery for each contract, they do not restrict the maximum number of impressions a contract can receive. As a result, high-priority contracts may monopolize premium exposure, leading to unfair and imbalanced allocations. \textbf{(3) Redundant exposures on the same page:} Request-level exclusivity fails to prevent the same contract from appearing in multiple slots within a single page, causing duplicate impressions and degraded user experience.

To address the aforementioned limitations, we propose an offline bipartite matching-based joint optimization framework tailored for multi-slot GD allocation. Our method fundamentally moves beyond the conventional single-slot paradigm by modeling the allocation between ad contracts and multiple ad slots at the page view level through a global offline optimization process, enabling coordinated decisions that enhance balance, fairness, and diversity beyond online greedy approaches.
We further introduce Page View (PV) constraints to cap per-slot impressions, preventing head-slot overuse and promoting balanced traffic allocation. In addition, we incorporate a Contract Roulette–based exclusivity mechanism that ensures that each contract appears in at most one slot per page, reducing redundant exposures and improving user experience. Our key contributions are as follows:
\begin{itemize}[leftmargin=*]
    \item We propose a unified joint optimization framework that formulates the contract-slot allocation problem as an offline bipartite matching task at the page view level, achieving globally coordinated and efficient fulfillment.
    \item We introduce two practical modules for industrial deployment: (i) a PV constraints for fine-grained traffic balance, and (ii) a Contract Roulette-Based Selection Mechanism for one-to-many assignment conflicts and reducing redundant exposures.
    \item Our proposed framework has been deployed in the GD advertising system of Meituan. Extensive online A/B testing validates its effectiveness and efficiency, where the Average Revenue Per User (ARPU) increased by \textbf{28.17\%}, and the Fulfillment Rate improved by \textbf{2.12\%} compared with the previous production baseline.
\end{itemize}

%To the best of our knowledge, this is the first to introduce per-slot PV constraints in contract-level allocation models.

\section{Related Work}
Research on GD Advertising has evolved significantly across both modeling objectives and solution strategies. Early approaches focused on fulfilling contractual guarantees through offline optimization. SHALE~\cite{bharadwaj2012shale} formulates GD advertising as a scalable quadratic program to meet impression delivery targets, while the dual-based method~\cite{chen2011real} further considers advertiser-side utility, bridging guarantee fulfillment and business effectiveness. Nearline control strategies were introduced to align delivery with real-time platform dynamics better. XShale~\cite{fang2019large} adjusts delivery pacing based on short-term feedback, optimizing advertiser outcomes. RAP~\cite{zhang2020request} further incorporates platform-level concerns such as traffic efficiency and delivery fairness, marking an early attempt toward multi-objective coordination in nearline settings. Recent work has formulated GDA as a sequential decision problem under complex constraints. AUAF~\cite{cheng2022adaptive} introduces slot-level mutual exclusivity in an optimal control framework, improving contract delivery precision. CONFLUX~\cite{wang2022conflux} and FACC~\cite{dai2023fairness} jointly optimize delivery, advertiser utility, and platform-wide objectives using multi-stage control mechanisms.

Despite their effectiveness, existing methods lack fine-grained page-view-level traffic control and explicit modeling of multi-slot coordination, limiting their ability to prevent overexposure and ensure fair delivery across diverse ad positions.

\begin{figure*}
    \centering
    \includegraphics[width=1.0\linewidth]{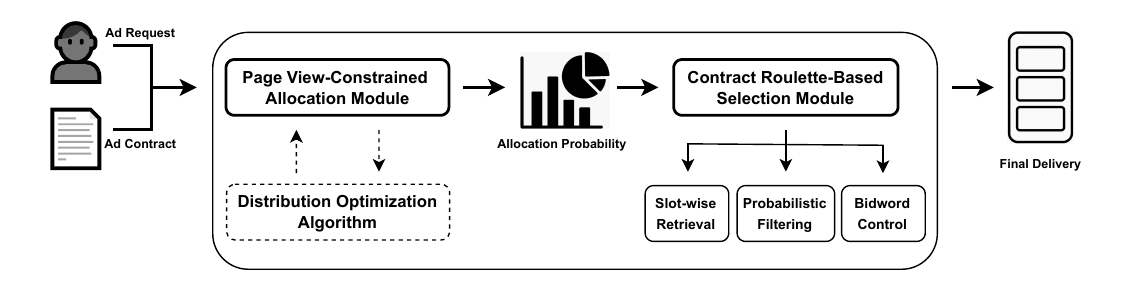}
    \caption{Overview of proposed framework.}
    \Description{The figure provides a visual summary of the proposed method.}
    \label{fig: overview}
\end{figure*}

\section{Methodology}
We formulate the multi-slot ad allocation problem as an offline bipartite matching task between ad requests and contracts, where each page view spans multiple slot positions. Our proposed framework, as illustrated in Figure~\ref{fig: overview}, consists of two key components: (1) Page View-Constrained Allocation  Module, which computes globally coordinated allocation probabilities while enforcing a novel Page View constraint to limit per-slot exposure and balance traffic across positions; and (2) Contract Roulette-Based Selection Module, which resolves one-to-many matching conflicts by probabilistically selecting a winning contract per request. %A key innovation of our framework is the introduction of a Page View constraint, which serves as a supply-side mechanism to limit the maximum exposure per slot. This constraint ensures that allocation decisions respect per-position impression budgets, mitigating over-concentration on premium slots and improving overall system robustness.

\subsection{Page View-Constrained Allocation Module}

To coordinate ad delivery across multiple slots within the same page view, we formulate the allocation task as a constrained optimization problem. This formulation captures not only contract fulfillment and user engagement but also fine-grained control over slot-level exposure. In particular, we introduce a novel Page View constraint to limit the number of impressions each slot can serve, thereby preventing over-exposure of high-traffic positions and promoting a more balanced traffic allocation across all ad slots. The detailed objective function and constraints are defined as follows, which consists of three parts. The first term is a smoothness regularization term that penalizes the deviation between the actual allocation $x_{ij}$ and the normalized target delivery ratio $\theta_j$, which helps stabilize contract delivery and improve allocation fairness. The second term is a priority-aware reward term that encourages the system to allocate more impressions to higher-priority contracts. The third term is an interest-aware matching term that promotes assignments with higher estimated user-interest relevance.
\begin{align}
\arg\min_{x_{ij}} \quad & 
\frac{1}{2} \sum_{j} \sum_{i \in \Gamma(j)} s_i \frac{V_j}{\theta_j} (x_{ij} - \theta_j)^2 
- \sum_{j} w_j \sum_{i \in \Gamma(j)} s_i x_{ij} \\
& - \sum_{j} \lambda_j \sum_{i \in \Gamma(j)} s_i x_{ij} c_{ij} \nonumber \\
\text{s.t.} \quad & 
\sum_{i \in \Gamma(j)} s_i x_{ij} \le d_j, \quad \forall j \\
& \sum_{j \in \Gamma(i)} x_{ij} \le 1, \quad \forall i \\
& x_{ij} \ge 0, \quad \forall i, j\\
& s_ix_{ij} \le pv_i, \quad \forall i, j
\end{align}
where $x_{ij}$ denotes the allocation probability from request $i$ (i.e., supply) to contract $j$ (i.e., demand), and $c_{ij}$ represents the user interest between them; $s_i$ denotes the capacity of request $i$, and $d_j$ represents the required number of impressions for contract $j$; $\lambda_j$ reflects the importance of user interest for demand $j$. Distinct from prior work, we introduce $pv_i$ to denote the page-view constraint of request $i$, enabling explicit control of slot-level exposure and balanced traffic allocation, which constitutes a key novelty of our formulation. Let $\Gamma(j)$ denote the set of requests that can serve contract $j$, and let $\Gamma(i)$ denote the set of contracts that request $i$ is eligible to serve.

\begin{align}
L(\alpha, \beta, \gamma, \delta) = & \frac{1}{2} \sum_{j} \sum_{i \in \Gamma(j)} s_i \frac{V_j}{\theta_j} (x_{ij} - \theta_j)^2 - \sum_{j} w_j \sum_{i \in \Gamma(j)} s_i x_{ij} \nonumber \\
&
- \sum_{j} \lambda_j \sum_{i \in \Gamma(j)} s_i x_{ij} c_{ij} \nonumber + \sum_{j} \alpha_j \left( \sum_{i \in \Gamma(j)} s_i x_{ij} - d_j \right) \\
& + \sum_{i} \beta_i \left( \sum_{j \in \Gamma(i)} x_{ij} - 1 \right) 
- \sum_{j} \sum_{i \in \Gamma(j)} \gamma_{ij} x_{ij} \nonumber \\
& + \sum_{j} \sum_{i \in \Gamma(j)} \delta_{ij} \left( s_i x_{ij} - pv_i \right)
\end{align}

\noindent \textbf{KKT conditions}:
\begin{equation}
s_i \frac{V_j}{\theta_j} (x_{ij} - \theta_j) - w_j s_i - \lambda_j s_i c_{ij} + \alpha_j s_i + \beta_i s_i - \gamma_{ij} + \delta_{ij} s_i = 0
\end{equation}

\begin{equation}
\alpha_j \left( \sum_{i \in \Gamma(j)} s_i x_{ij} - d_j \right) = 0
\label{equ: alpha}
\end{equation}

\begin{equation}
\beta_i \left( \sum_{j \in \Gamma(i)} x_{ij} - 1 \right) = 0
\label{equ: beta}
\end{equation}

\begin{equation}
\gamma_{ij} x_{ij} = 0
\label{equ: gamma}
\end{equation}

\begin{equation}
\delta_{ij} \left( s_i x_{ij} - pv_i \right) = 0
\label{equ: delta}
\end{equation}

\begin{equation}
\alpha_j \geq 0,\quad \beta_i \geq 0,\quad \gamma_{ij} \geq 0,\quad \delta_{ij} \geq 0 
\end{equation}

\noindent where $\alpha_j$, $\beta_i$, $\gamma_{ij}$ and $\delta_{ij}$ are the Lagrangian multipliers of constraints
Equation \ref{equ: alpha}, \ref{equ: beta}, \ref{equ: gamma} and \ref{equ: delta} respectively. According to the KKT conditions, the optimal allocation probability $x_{ij}$ can be derived as:
\begin{equation}
x_{ij} = \max\left\{ 0,\ \theta_j \left(1 + \frac{w_j + \lambda_j c_{ij} - \alpha_j - \beta_i - \delta_{ij}}{V_j} \right) \right\}
\end{equation}

To efficiently solve the PV-constrained allocation problem, we adopt an online distributed strategy that updates the dual variables $\alpha_j$, $\beta_i$, and $\delta_{ij}$ based on the KKT conditions. These dual variables correspond to the contract demand, request capacity, and slot-level page view constraints, respectively. Instead of solving the primal problem directly, which is computationally expensive in large-scale settings, we leverage the closed-form expression of $x_{ij}$ derived from the KKT optimality condition and iteratively adjust the dual variables using projected gradient steps.

At each iteration, $\alpha_j$ is increased if the total allocated impressions to contract $j$ exceed its demand $d_j$, while $\beta_i$ is updated to ensure the total allocation from request $i$ remains within its capacity. Similarly, $\delta_{ij}$ is adjusted to enforce the fine-grained page view constraint for each ad slot. Each contract $j$ is characterized by a priority weight $w_j$, a smoothness parameter $V_j$, and a normalized target delivery ratio $\theta_j = \frac{d_j}{\sum_{i \in \Gamma(j)} s_i}$, where $V_j$ controls the strength of the fairness regularization for the $j$-th contract. The gradient of the objective with respect to $x_{ij}$ is used to guide the iterative updates. These operations are executed in parallel across all $(i, j)$ pairs, ensuring scalability and real-time adaptability. The overall procedure is summarized in Algorithm~\ref{algorithm:test}.

\begin{algorithm}
\caption{PV Constrained Allocation Algorithm}
\label{algorithm:test}
\LinesNumbered
\KwIn {$s_i, d_j, \lambda_j, V_j, w_j, c_{ij}$}
\KwOut {$\alpha_{j}, \beta_{i}, \delta_{ij}, x_{ij}$}
Step 1: Initialize. Set $\alpha_j=w_j+\lambda_jc_{ij}, \theta_j $, calculate $x_{ij}$, and gradient $grad_j$ \\
Step 2:

\For{iteration = 1 to n}{
    update $\alpha_j$ with Equation: $\alpha_j^{t+1} = \alpha_j^t - V_j(1-\frac{d_j(\alpha^t)}{d_j})$ \\
    solve $\beta_i$: $\beta_i \leftarrow \max\left(0, \beta_i + \eta_\beta \left( \sum_{j \in \Gamma(i)} x_{ij} - 1 \right) \right)$ \\
    calculate $\delta_{ij}$: $\delta_{ij} \leftarrow \max\left(0, \delta_{ij} + \eta_\delta \left( s_i x_{ij} - pv_i \right) \right)$ \\
    calculate $x_{ij}$: $x_{ij} = \max\left\{ 0,\ \theta_j \left(1 + \frac{w_j + \lambda_j c_{ij} - \alpha_j - \beta_i - \delta_{ij}}{V_j} \right) \right\}$ \\
    update $grad_j$: $grad_j = \sum_{i \in \Gamma(j)} s_i x_{ij} - d_j$
}
\end{algorithm}

\subsection{Contract Roulette-Based Selection Module}

To ensure that each contract advertisement appears at most once per page and to maximize delivery diversity across ad slots, we design a contract roulette-based selection mechanism that integrates efficient candidate generation, probabilistic filtering, and adaptive bidword control under online latency constraints.

\subsubsection{Slot-wise Retrieval.}
In the recall phase, each ad slot $s_i$ independently retrieves a set of candidate contracts $\mathcal{A}_{s_i}$ based on the current query $q$ and an associated bidword $b_i$. To maintain fairness and long-tail exposure, a roulette sampling strategy is applied. For each candidate contract $a \in \mathcal{A}_{s_i}$, the probability of being recalled is proportional to its delivery weight $w(a)$:
\begin{equation}
    P(a \mid s_i) = \frac{w(a)}{\sum_{a' \in \mathcal{A}_{s_i}} w(a')}
\end{equation}
This design prioritizes contracts with higher delivery urgency while remaining compatible with budget delivery requirements and real-time demand-supply conditions. It also enforces that each POI-level contract can be recalled for only one slot per page, thus preventing redundant exposures and ensuring slot-level diversity.

\subsubsection{Probabilistic Filtering.} %and Position-Aware De-duplication.}
Once candidates are recalled, a multi-stage filtering process is applied. First, contracts are scored within each slot using a utility function that reflects both click-through rate and post-click conversion rate:

\begin{equation}
    \text{Score}(a) = \text{CTR}_a \cdot \text{CVR}_a
\end{equation}
Candidates with $\text{Score}(a) \geq K_1$ are retained, where $K_1$ is a system-defined threshold. For contracts appearing in multiple slots, only the one with the best relative position is preserved:

\begin{equation}
    s^*(a) = \arg\min_{s_i} \text{Rank}_{s_i}(a)
\end{equation}
where $\text{Rank}_{s_i}(a)$ denotes the intra-slot rank of contract $a$ in slot $s_i$. This ensures that each contract is bound to at most one slot, prioritizing the position where it has the strongest competitive advantage.

\begin{table*}[ht]
\centering
\caption{35\% and 70\% Gray-Scale Experiment Results}
\resizebox{1.0\textwidth}{!}{
\begin{tabular}{clcccccc}
\toprule
\textbf{Gray Scale} &
\multicolumn{1}{c}{\textbf{Experiment}} &
\multicolumn{4}{c}{\textbf{Merchant Efficiency}} &
\textbf{Platform Revenue} & \textbf{Contract Fulfillment} \\
\cmidrule(lr){3-6} \cmidrule(lr){7-7} \cmidrule(lr){8-8}
& & \textbf{Merchant ROI} & \textbf{Payment ROI} & \textbf{CTR} & \textbf{Payment CVR}
& \textbf{ARPU} & \textbf{Fulfillment Rate} \\
\midrule

\multirow{4}{*}{35\%}
 & A/A Effect (Treatment) & -33.98\% & 79.58\% & 17.78\% & 25.62\% & 0.71\% & 0.22\% \\
 & A/A Effect (Control)   & -38.63\% & 38.52\% & 0.58\%  & -5.85\% & 31.48\% & 1.32\% \\
 & A/B Effect             & -31.73\% & 54.96\% & 26.33\% & 32.23\% & -5.95\% & -10.39\% \\
 & DID Effect             & 4.65\%   & 41.06\% & 17.20\% & 31.47\% & -30.77\% & -11.50\% \\
\midrule
\multirow{4}{*}{70\%}
 & A/A Effect (Treatment) & 25.06\% & -4.75\% & 12.10\% & 13.35\% & 49.87\% & -2.90\% \\
 & A/A Effect (Control)   & -17.11\% & -33.87\% & 4.42\% & -10.01\% & 21.70\% & -5.03\% \\
 & A/B Effect             & -41.82\% & 64.76\% & 0.99\% & 4.57\% & 28.99\% & -3.95\% \\
 & DID Effect             & 42.17\% & 29.13\% & 7.67\% & 23.35\% & 28.17\% & 2.12\% \\
\bottomrule
\end{tabular}
}
\label{tab:gray}
\end{table*}

\subsubsection{Adaptive Bidword Control.}
To support dynamic retrieval across slots, we design an adaptive bidword selection strategy that ensures each slot retrieves relevant candidates while promoting diversity and contextual consistency across the page. For each slot $s_i$, the selected bidword $b_i$ is determined as:

\begin{equation}
\begin{aligned}
& b_i =
\begin{cases}
\text{RandomSample}(\mathcal{B}_q), & \text{if } i = 1 \\[0.5ex]
\text{RandomSample}(\mathcal{B}_q^{\text{avail}}(i)), & \text{if } i > 1 \text{ and } \mathcal{A}_{s_i}(b_{i-1}) = \emptyset \\[0.5ex]
b_{i-1}, & \text{if } i > 1 \text{ and } \mathcal{A}_{s_i}(b_{i-1}) \neq \emptyset
\end{cases}
\end{aligned}
\end{equation}
where $\mathcal{B}_q$ denotes the set of candidate bidwords for query $q$, and $\mathcal{A}_{s_i}(b)$ is the set of ads retrievable for slot $s_i$ using bidword $b$. The available bidwords for slot $s_i$ are defined as:
\begin{equation}
\mathcal{B}_q^{\text{avail}}(i) = \left\{ b \in \mathcal{B}_q \mid \mathcal{A}_{s_i}(b) \neq \emptyset \right\} \setminus \{b_1, \dots, b_{i-1}\}
\end{equation}

When determining the bidword for the first slot, the system randomly samples from the full bidword pool $\mathcal{B}_q$. Subsequent slots preferentially reuse the previous bidword $b_{i-1}$ if it can still retrieve valid ads; otherwise, the system switches to another eligible bidword from the remaining available pool. In practice, this fallback process is also guided by inventory availability and delivery deficit, especially when certain bidwords are highly supply-constrained. This design ensures relevant ads for each slot while reducing redundancy and maintaining page-level coherence.

\section{Experiments}
To evaluate the real-world effectiveness of our proposed multi-slot allocation framework, we conducted two rounds of online gray-scale experiments on the Meituan advertising platform, where our method was fully deployed and actively served real traffic. Then we discuss experimental results.

\subsection{Experimental Settings}
The experiments were conducted on Meituan’s production environment with real online traffic and real-time ad delivery, under 35\% and 70\% gray-scale settings, where the treatment and control groups were split by corresponding proportions of POIs. Both experiments were evaluated using a combination of the A/A test to validate group-level stability, the A/B test for one-day cross-group comparison, and the DID (Difference-in-Differences) analysis to estimate the net causal effect while controlling for temporal fluctuations. The online experiment followed a progressive gray-scale rollout.  For the 35\% gray-scale setting, the baseline period was from March 29 to April 2, 2025, and the experimental period was from April 3 to April 7, 2025. For the 70\% gray-scale setting, the baseline period was from March 27 to April 1, 2025, and the experimental period was from April 9 to April 14, 2025.

%To assess the effectiveness of our proposed method, we evaluate the system from three aspects: merchant efficiency, platform revenue, and contract fulfillment to capture merchant-side benefits and platform-level business objectives jointly. Merchant efficiency includes Merchant Return on Investment (ROI), Payment Return on Investment (ROI), Click-Through Rate (CTR), and Payment Conversion Rate (CVR), which reflect user engagement and conversion quality. Platform revenue is measured by Average Revenue Per User (ARPU). Contract fulfillment is measured by fulfillment Rate, which reflects the delivery stability of guaranteed contracts.

\subsection{Evaluation Metrics}

To assess the effectiveness of our proposed method, we adopt a comprehensive set of evaluation metrics, grouped into three key dimensions: \textit{Merchant Efficiency}, \textit{Platform Revenue}, and \textit{Contract Fulfillment} to capture merchant-side benefits and platform-level business objectives jointly:
\begin{itemize}[leftmargin=*]
    \item \textbf{Merchant Return on Investment (ROI)}: Measures the return on investment from the merchant’s perspective, computed as revenue generated through advertising divided by ad spend.
    \item \textbf{Payment Return on Investment (ROI)}: Focuses on actual purchase behaviour, providing a more direct indicator of commercial effectiveness.
    \item \textbf{Click-Through Rate (CTR)}: Captures user engagement with advertisements, indicating exposure quality.
    \item \textbf{Payment Conversion Rate (CVR)}: The proportion of clicks that result in a completed payment, reflecting the efficacy of the ad content and targeting.
    \item \textbf{ARPU Average Revenue Per User (ARPU)}: Reflects monetization efficiency per user, capturing per-capita revenue yield.
    \item \textbf{Fulfillment Rate}: The ratio of fulfilled to planned contractual objectives to evaluate service quality and delivery reliability.
\end{itemize}

\subsection{Performance Analysis}
%As shown in Tables \ref{tab:gray}, our method consistently outperforms across key dimensions, merchant efficiency, platform monetization, and contractual delivery quality. Our proposed method yields substantial gains in merchant-side metrics. In particular, the DID results demonstrate a 42.17\% improvement in Merchant ROI, 29.13\% in Payment ROI, 7.67\% in CTR, and 23.35\% in Payment CVR, indicating enhanced user engagement and stronger conversion performance throughout the advertising funnel. These improvements highlight the method’s effectiveness in not only attracting user attention but also driving purchase behaviours. On the platform side, ARPU increases by 28.17\%, suggesting better monetization efficiency per user under the multi-slot optimization strategy. While some volatility may still exist in secondary metrics, the overall ARPU trend indicates improved budget utilization and delivery prioritization. In terms of contract fulfillment, the Fulfillment Rate shows a positive DID effect of 2.12\%, suggesting that the proposed allocation framework maintains stable and reliable contract delivery even under expanded exposure. %This further confirms the robustness of the system under scaled deployment.

%In summary, the results demonstrate that our joint optimization approach consistently improves merchant returns, user interaction quality, and revenue efficiency while maintaining contract stabilityand scalability across different traffic levels, demonstrating strong practical value in real-world advertising deployments.

We conduct controlled A/A and A/B tests under two gray-scale settings (35\% and 70\%) to rigorously evaluate the effectiveness and robustness of our method. As shown in Tables \ref{tab:gray}, our method consistently outperforms across key dimensions, merchant efficiency, platform monetization, and contractual delivery quality.

Our proposed method yields substantial gains in merchant-side metrics. In particular, the DID (Difference-in-Differences) results demonstrate a 42.17\% improvement in Merchant ROI, 29.13\% in Payment ROI, 7.67\% in CTR, and 23.35\% in Payment CVR, indicating enhanced user engagement and stronger conversion performance throughout the advertising funnel. These improvements highlight the method’s effectiveness in not only attracting user attention but also driving purchase behaviours.

On the platform side, ARPU increases by 28.17\%, suggesting better monetization efficiency per user under the multi-slot optimization strategy. While some volatility may still exist in secondary metrics, the overall ARPU trend indicates improved budget utilization and delivery prioritization.

In terms of contract fulfillment, the Fulfillment Rate shows a positive DID effect of 2.12\%, suggesting that the proposed allocation framework maintains stable and reliable contract delivery even under expanded exposure. This further confirms the robustness of the system under scaled deployment.

In summary, the experimental results demonstrate that our approach consistently improves merchant return, user interaction quality, and revenue efficiency while maintaining contract integrity and scalability across different traffic levels, demonstrating strong practical value in real-world advertising deployments.

\section{Conclusion}
This paper presents a unified framework for guaranteed display advertising in multi-slot environments. By modeling the contract-slot assignment as a page-view-level bipartite matching problem, and introducing novel constraints and selection mechanisms, our approach achieves fine-grained traffic control, eliminates redundant exposures, and enhances fairness in ad delivery. The proposed Page View-constrained allocation model enables precise regulation of slot-level traffic, while the contract roulette mechanism ensures diversity and mutual exclusivity. Extensive online experiments on Meituan validate the effectiveness of our framework across key metrics. These results underscore the practical value of coordinated multi-slot optimization in industrial GD systems and provide a foundation for future research on fairness-aware and scalable ad delivery mechanisms.

%\begin{acks}
%\end{acks}

\bibliographystyle{ACM-Reference-Format}
\bibliography{sample-base}

@inproceedings{cheng2022adaptive,
  title={An adaptive unified allocation framework for guaranteed display advertising},
  author={Cheng, Xiao and Liu, Chuanren and Dai, Liang and Zhang, Peng and Fang, Zhen and Zu, Zhonglin},
  booktitle={Proceedings of the Fifteenth ACM International Conference on Web Search and Data Mining},
  pages={132--140},
  year={2022}
}

@inproceedings{dai2024percentile,
  title={Percentile risk-constrained budget pacing for guaranteed display advertising in online optimization},
  author={Dai, Liang and Lyu, Kejie and Zhang, Chengcheng and Zhao, Guangming and Zu, Zhonglin and Wang, Liang and Zheng, Bo},
  booktitle={Proceedings of the AAAI Conference on Artificial Intelligence},
  volume={38},
  number={8},
  pages={7987--7994},
  year={2024}
}

@inproceedings{fang2019large,
  title={Large-scale personalized delivery for guaranteed display advertising with real-time pacing},
  author={Fang, Zhen and Li, Yang and Liu, Chuanren and Zhu, Wenxiang and Zheng, Yu and Zhou, Wenjun},
  booktitle={2019 IEEE International Conference on Data Mining (ICDM)},
  pages={190--199},
  year={2019},
  organization={IEEE}
}

@inproceedings{hojjat2014delivering,
  title={Delivering guaranteed display ads under reach and frequency requirements},
  author={Hojjat, Ali and Turner, John and Cetintas, Suleyman and Yang, Jian},
  booktitle={Proceedings of the AAAI Conference on Artificial Intelligence},
  volume={28},
  number={1},
  year={2014}
}

@inproceedings{lei2020multi,
  title={Multi-objective optimization for guaranteed delivery in video service platform},
  author={Lei, Hang and Zhao, Yin and Cai, Longjun},
  booktitle={Proceedings of the 26th ACM SIGKDD International Conference on Knowledge Discovery \& Data Mining},
  pages={3017--3025},
  year={2020}
}

@inproceedings{bharadwaj2012shale,
  title={Shale: an efficient algorithm for allocation of guaranteed display advertising},
  author={Bharadwaj, Vijay and Chen, Peiji and Ma, Wenjing and Nagarajan, Chandrashekhar and Tomlin, John and Vassilvitskii, Sergei and Vee, Erik and Yang, Jian},
  booktitle={Proceedings of the 18th ACM SIGKDD international conference on Knowledge discovery and data mining},
  pages={1195--1203},
  year={2012}
}

@inproceedings{zhang2020request,
  title={A request-level guaranteed delivery advertising planning: Forecasting and allocation},
  author={Zhang, Hong and Zhang, Lan and Xu, Lan and Ma, Xiaoyang and Wu, Zhengtao and Tang, Cong and Xu, Wei and Yang, Yiguo},
  booktitle={Proceedings of the 26th ACM SIGKDD International Conference on Knowledge Discovery \& Data Mining},
  pages={2980--2988},
  year={2020}
}

@inproceedings{dai2023fairness,
  title={Fairness-aware guaranteed display advertising allocation under traffic cost constraint},
  author={Dai, Liang and Zu, Zhonglin and Wu, Hao and Wang, Liang and Zheng, Bo},
  booktitle={Proceedings of the ACM Web Conference 2023},
  pages={3572--3580},
  year={2023}
}

@inproceedings{wang2022conflux,
  title={CONFLUX: A Request-level Fusion Framework for Impression Allocation via Cascade Distillation},
  author={Wang, XiaoYu and Tan, Bin and Guo, Yonghui and Yang, Tao and Huang, Dongbo and Xu, Lan and Freris, Nikolaos M and Zhou, Hao and Li, Xiang-Yang},
  booktitle={Proceedings of the 28th ACM SIGKDD Conference on Knowledge Discovery and Data Mining},
  pages={4070--4078},
  year={2022}
}

@inproceedings{chen2011real,
  title={Real-time bidding algorithms for performance-based display ad allocation},
  author={Chen, Ye and Berkhin, Pavel and Anderson, Bo and Devanur, Nikhil R},
  booktitle={Proceedings of the 17th ACM SIGKDD international conference on Knowledge discovery and data mining},
  pages={1307--1315},
  year={2011}
}

@inproceedings{li2024bi,
  title={Bi-Objective Contract Allocation for Guaranteed Delivery Advertising},
  author={Li, Yan and Huang, Yundu and Mao, Wuyang and Ye, Furong and He, Xiang and Zu, Zhonglin and Cai, Shaowei},
  booktitle={Proceedings of the 30th ACM SIGKDD Conference on Knowledge Discovery and Data Mining},
  pages={1691--1700},
  year={2024}
}

@inproceedings{mao2023end,
  title={End-to-End Inventory Prediction and Contract Allocation for Guaranteed Delivery Advertising},
  author={Mao, Wuyang and Liu, Chuanren and Huang, Yundu and Zu, Zhonglin and Harshvardhan, M and Wang, Liang and Zheng, Bo},
  booktitle={Proceedings of the 29th ACM SIGKDD Conference on Knowledge Discovery and Data Mining},
  pages={1677--1686},
  year={2023}
}

@article{lei2025generative,
  title={Generative Large-Scale Pre-trained Models for Automated Ad Bidding Optimization},
  author={Lei, Yu and Zhao, Jiayang and Zhao, Yilei and Zhang, Zhaoqi and Cai, Linyou and Xie, Qianlong and Wang, Xingxing},
  journal={arXiv preprint arXiv:2508.02002},
  year={2025}
}

\appendix

\end{document}